# Cause vs. Effect in Context-Sensitive Prediction of Business Process Instances[*]

Jens Brunk[a,*], Matthias Stierle[b], Leon Papke[a], Kate Revoredo[c], Martin Matzner[b], Jörg Becker[a]

[a]*University of Muenster - ERCIS, Münster, Germany*
[b]*Institute of Information Systems, Friedrich-Alexander Universität Nürnberg-Erlangen, Germany*
[c]*Department of Information Systems and Operations, Vienna University of Economics and Business (WU), Vienna, Austria*

**Abstract**

Predicting undesirable events during the execution of a business process instance provides the process participants with an opportunity to intervene and keep the process aligned with its goals. Few approaches for tackling this challenge consider a multi-perspective view, where the flow perspective of the process is combined with its surrounding context. Given the many sources of data in today's world, context can vary widely and have various meanings. This paper addresses the issue of context being cause or effect of the next event and its impact on next event prediction. We leverage previous work on probabilistic models to develop a Dynamic Bayesian Network technique. Probabilistic models are considered comprehensible and they allow the end-user and his or her understanding of the domain to be involved in the prediction. Our technique models context attributes that have either a cause or effect relationship towards the event. We evaluate our technique with two real-life data sets and benchmark it with other techniques from the field of predictive process monitoring. The results show that our solution achieves superior prediction results if context information is correctly introduced into the model.

*Keywords:* predictive process monitoring, process prediction, context-sensitivity, dynamic bayesian network, process mining

## 1. Introduction

In the pursuit of gaining control over their business processes, organizations seek ways to manage their business processes proactively. *Process mining*, which has emerged as a technology that allows process analysts to discover, enhance, and check the conformance of business processes using data (Van Der Aalst, 2016), presents an excellent toolbox with which organizations can create transparency. One branch of process mining is predictive business process monitoring, which deals with techniques and methods to predict process behavior, such as the next process step, throughput times, or risk assessments (Maggi et al., 2014). Possible use cases include cases of suspected fraudulent process behavior or the need to suggest actions about the further processing of a process instance.

In the field of predictive process monitoring (PPM), a multitude of different techniques, most of which seek to improve prediction accuracy, have been proposed in recent years. Márquez-Chamorro et al. (2018) and Di Francesco-marino et al. (2018) classified these techniques by the input data, the algorithm type, and the prediction target. Tama and Comuzzi (2019) provide an empirical comparison of different classification techniques for next event prediction. Especially for techniques that predict the next process step(s), neural networks have been a popular choice, as they have shown promising results in other fields (Hinkka et al., 2018; Schönig et al., 2018; Khan et al., 2018; Park and Song, 2019). However, a well-known drawback of neural networks is their *black box* design, which makes it difficult to understand why a particular prediction was made (Malioutov et al., 2017). Previous research has shown that users are skeptical about decision support systems if they do not understand it (Martens and Provost, 2014; Gregor and Benbasat, 1999). Therefore, to apply predictive business process monitoring in an organization, the system should be comprehensible and provide meaningful insights (Verenich et al., 2019).

Breuker et al. (2016) proposed probabilistic modeling techniques for business process prediction, as they provide comprehensible process models that users can easily understand. In their recent work, Park and Song (2019) point out the importance of reliable prediction results for decision making of managers, which can be supported by quantifying the prediction uncertainty based on probability distributions. Probabilistic techniques can also deal with noise – a common issue in process data (i.e., event logs) (Van Der Aalst, 2016).

The suitability of probabilistic techniques for process prediction is supported by the results of Breuker et al. (2016), although their experiments revealed room for improvement with regard to their prediction accuracy in large part because the technique considers only the control-flow information present in the event log. Taking the example of a credit loan application process, one would, however, assume that the loan amount, the customer's risk rating, and even the employee who processes the application all have significant impacts on how the application is further processed. As a result, many techniques use additional context information that is present in the event log (Senderovich et al., 2019; Verenich et al., 2019; Camargo et al., 2019) or even include data from external sources like sensors (Borkowski et al., 2019) and news (Yeshchenko et al., 2018).

We extend the work of Breuker et al. (2016) by considering the context. As a probabilistic model, we use a manually constructed dynamic Bayesian network (DBN) that allows the representation of context.

We propose a structure that considers the impacts that a context variable can have on a running process instance. Therefore, we adopt the concept of symptom and background variables, which was suggested by Kjaerulff and Madsen (2007), for event logs. We distinguish between context variables that affect the current process step (background) and context variables affected by the current process step (symptom). Therefore, the objective of our research is to:

**Research Goal** Develop a context-sensitive prediction technique for business processes that considers cause and effect relationships among an event log's variables.

The main contribution of this paper is a technique based on DBNs that is capable of context-sensitive process prediction. Our artifact contributes to the discussion on how to use event log data and its associated contexts for prediction by introducing the concept of symptom and background variables. By instantiating our technique, we also contribute to practice by offering the possibility to make use of the artifact for real-world data sets.

Our approach is in line with the design science research (DSR) paradigm (Hevner, 2004). We develop a dynamic Bayesian network structure as an artifact and instantiate it to evaluate its efficacy with data, following the DSR process suggested by Peffers et al. (2007). Our research uses a problem-centered technique, as it commences with the *Identify & Motivate Problem* phase.

This introduction demonstrates the importance of context-sensitive predictive process monitoring and why it is essential to develop tools to evaluate it further. In section 2, we recapitulate on the current state-of-the-art of PPM, we introduce the concept of Bayesian networks (BNs), and we discuss the importance of comprehensibility of predictive methods. A fundamental construct of the DSR process is continual iterations to improve the developed artifact. A first iteration of the prediction technique was previously presented and discussed in an international symposium on Data-driven Process Discovery and Analysis (SIMPDA) (Brunk et al., 2018). The gathered feedback led to our reiteration of the *Design & Development* phase to exploit unused potential, thus improving the technique. This paper focuses on the second iteration of the process. In section 3, we describe the *Design & Development* phase and sections 5 and 6 present the *Demonstration and Evaluation* of our technique.

## 2. Research Background

### 2.1. Predictive Process Monitoring

The business process management (BPM) research field is concerned with the discovery, analysis, redesign, and monitoring of business processes with the ultimate goal of improving them (Dumas et al., 2013). For many years, BPM focused on modeling the to-be processes that depict the ideal process flow.

However, the availability of large volumes of transactional data from information systems that support business processes has shifted this focus towards analytics of the as-is process. As a result, process mining emerged as a new research field that deals with techniques and methods based on event logs, which are data logs that contain the history of a process instance and that are used for discovery, conformance checking, and enhancement of business processes (Van Der Aalst, 2016).

With increased activity in predictive analytics and machine learning (ML), a new branch of process mining called PPM was created that deals with predicting future process behavior to improve process performance and minimize risks on an operational level (Márquez-Chamorro et al., 2018). As such, PPM techniques are expected to deliver insights into the future development of a process instance at runtime (Di Francescomarino et al., 2018).

A central differentiator between predictive techniques in general and techniques designed especially for PPM is that the latter revolves around the concept of *process awareness*, which indicates that the predictive model "exploits an explicit representation of the process model to make the prediction" (Márquez-Chamorro et al., 2018, p.3). In other words, the model accounts for the fact that an event log consists of one to many cases, and each case has one to many events that relate to each other chronologically.

Process awareness is commonly achieved by transforming the event log into a suitable data structure, referred to as *encoding* (Márquez-Chamorro et al., 2018), or by selecting a predictive method that adequately considers the temporal dependency of the events within a process instance.

Early techniques focused on the aspect of *process awareness* (e.g. Breuker et al. 2016; Tax et al. 2017) and considered only the control-flow dimension of the process as an explanatory variable. Given that a business process is usually executed in the same way each time it runs (the *happy path*), acceptably accurate results can often be achieved by simply suggesting the most frequent event or sequence as a prediction. For techniques to compete against these naive but computationally inexpensive approaches, they must take into account further information about the process and its environment. As a result, increasing numbers of techniques are also incorporating contextual information (Márquez-Chamorro et al., 2018) that can range from the resource used for executing a process step to the value of an invoice that is processed and even weather data. Contextual information can be both static during the execution of a process instance or dynamic, such that the attribute changes from one event to the next (Leontjeva et al., 2015).

The objective of PPM techniques is to predict the next event(s), process duration, or process outcomes like risk assessments (Márquez-Chamorro et al., 2018). We contribute to the research stream that deals with the prediction of the next event of a process instance.

Predicting the next event of a business process is a multi-class classification problem. Each possible next event represents one class that an algorithm can predict. How well such a (multi-class) classification method performs can be evaluated based on several measures. Which measure to choose and subse-quently to optimize depends on the objective of the prediction task. In PPM, the objective is commonly stated simply as *predicting the next event*. However, often no precise use case is given, which leads to the assumption that every event has equal relevance (e.g. in regards to importance, risk, cost, etc.). As such, *Accuracy* – which indicates the technique's overall performance by dividing the number of correct predictions by the total number of predictions (Sokolova and Lapalme, 2009) – is the most popular measure. For example, it was used by Tama and Comuzzi (2019) in their recent comparison of techniques predicting next events. *Accuracy* is an intuitive measure that allows a general statement to be made about a technique's predictive power. However, PPM is subject to class imbalances as certain events and event sequences occur much more frequently than others (Marquez-Chamorro et al., 2017). Accordingly, a high level of *Accuracy* can often be achieved by naive approaches like n-grams (Breuker et al., 2016), which basically counts the different variants of process executions and predicts based on the highest frequencies.

Therefore, other works also present results for the *F1-score*. It is computed as the harmonic mean of *Precision* and *Recall*. *Precision* can be used to determine the predictive quality of the method given that a specific class has been predicted. *Recall* (or Sensitivity), on the other hand, measures how well the method can select instances of a particular class from the data. Therefore, it puts more emphasis on false positives and negatives.

*2.2. Bayesian Networks*

One common family of predictive methods are BNs. BNs are used to construct a probabilistic model of the relationships in a complex system (Koller and Friedman, 2009, p. 2 f.). A complex system is characterized as a set of random variables, each of which describes a property of the real world. BNs help to suggest reasonable actions in these systems under conditions of uncertainty (Holmes and Jain, 2008; Korb and Nicholson, 2010). BNs use graphical models to represent their networks and the causal interactions among the sets of variables in them (Kjaerulff and Madsen, 2007, p. 17). A Directed Acyclic Graph (DAG) is such a graphical model used to represent probabilistic networks (Koller and Friedman, 2009). The nodes in a DAG represent the variables of a BN, and the edges between nodes indicate (conditional) dependence between the variables.

A business process is defined as a "timely and logical sequence of activities" (Becker et al., 2012, p.4), so to use BNs for process prediction, the model has to account for this change over time. DBNs are a class of BNs that are used to model dynamic systems like stochastic processes (Murphy, 2002, p. 14). DBNs belong to the category of temporal models, which reason about the state of the world as it evolves over time (Koller and Friedman, 2009, p. 200). The structure and parameters of a DBN are usually fixed, so they do not adjust over time. Therefore, a DBN can be seen as a network representation of a dynamic system (Murphy, 2012, p. 628 f.).

Following this understanding, a DBN is a compact representation generated from an infinite set of BNs, one for every time slice $t$. A DBN, as a BN, has two components, the structure and the probability distribution (i.e., parameters) associated with each variable (Koller and Friedman, 2009; Murphy, 2002). Both components can be specified manually or learned from data. The learning of the network structure from the data is a computationally expensive approach. Also, event logs suffer from noise, i.e., they contain "rare and infrequent behavior not representative for the typical behavior of the process" (van der Aalst, 2016, p. 185). While noise could potentially be reduced by filtering methods (e.g. Cheng and Kumar, 2015), it still might lead to highly complex network structures. Both the high learning time and the incomprehensible network structures might hinder adoption in practice.

The two main tasks when working with DBNs are *learning* of the parameters and *inference*. Parameter learning refers to estimating the values of parameters that correspond to the structure of a graph $G$ and the distributions $P$ of a BN from data.

The task of computing the posterior probability distribution for a set of nodes (random variables), given values for the nodes (evidence), and a joint distribution for the model is called inference. Inference is often also referred to as *belief updating/propagation* or *probabilistic inference* (Korb and Nicholson, 2010, p. 55). Kjaerulff (1995) introduced the first computational system for dynamic time-sliced BNs. Exact inference can be computationally expensive because all hidden variables can become correlated over time, a process called entanglement (Koller and Friedman, 2009, pp. 656–660). For this reason, real-world applications usually implement algorithms that perform approximate inferences. Inferences in DBNs can be achieved via filtering, prediction, and smoothing (Kjaerulff and Madsen, 2007, p. 99). Filtering is the process of extracting information about the state of the system at the current time slice, predicting is for determining the state of the system at a future time slice, and smoothing yields information about previous time slices given evidence about a current time slice.

In the context of PPM, Leontjeva et al. (2015) applied hidden Markov models (HMMs), a type of DBNs, for predicting process outcomes (i.e.,, not for predicting the next event). They trained a large number of models for each combination of attributes. Breuker et al. (2016) suggested the use of DBNs for next-event prediction because of their success in the field of grammatical inference and the analogy to processes and sentences or, rather, events and words. The authors advised against using HMMs, though, as the current observation does not impact the next observation, which is counter-intuitive for process events, and suggested a probabilistic finite automata (PFA) instead, as shown in Figure 1. Both HMMs and PFA are particular cases of a DBNs.

### 2.3. Comprehensibility of Predictive Methods

The importance of ML techniques' comprehensibility in certain scenarios was recognized some time ago. Shortliffe (1976) argued that, in medical applications, physicians have to be able to understand how a particular prediction was made if they are to decide about the patient's treatment. Martens and Provost (2014) found that the same applies to the business domain, where managers want to understand a prediction model before they put trust in it.

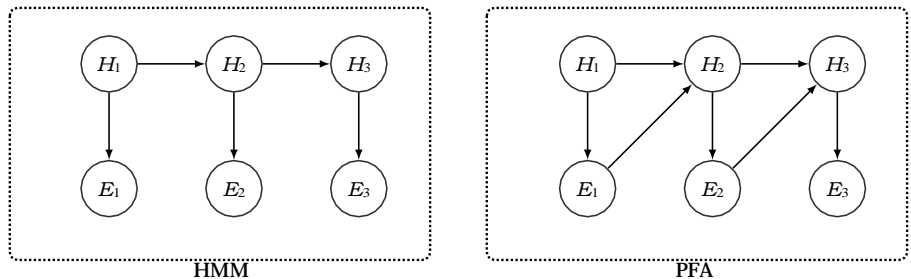

Figure 1: HMM vs. PFA network structure (H = hidden state, E = Event)

This stream of research was resurrected together with the renewed interest in artificial intelligence (AI) and is often referred to as *explainable artificial intelligence (XAI)* (Adadi and Berrada, 2018). It acknowledges the need for transparency in the decisions of a prediction algorithm to increase user acceptance (Miller, 2018). In their recent review of the field, Adadi and Berrada (2018) pointed out that there is no agreed definition of XAI, but several movements have sought to increase the transparency of the decisions ML algorithms make. This concept is usually referred to as explainable AI, interpretable AI, or comprehensible AI; this paper uses the last of these terms. As such, a ML algorithm should be comprehensible.

According to Du et al. (2019), comprehensibility of machine learning techniques can either be delivered by intrinsic models that are explainable by design (*ante-hoc*) or *post-hoc* by creating a second model that provides explanations. Furthermore, the explanations can either be made locally for single predictions or globally for the model (Du et al., 2019).

While an increasing amount of research is directed towards *post-hoc* explanations of deep learning techniques, the faithfulness of the explanations (i.e., correctly representing the model's decision process) derived from black-box models remains a challenge (Du et al., 2019). For DBNs, the conditional probability distributions (CPDs) of the trained model can be extracted for each possible time slice, state, and evidence input. Hence, the decisions of the network are transparent, and the CPD can be used to create both global and local post-hoc explanations (Du et al., 2019).

In the context of PPM, Park and Song (2019) discuss the importance of comprehensibility to support decision making. The probability of the prediction is mentioned as a way to improve confidence in it as a simple example for a local, post-hoc explanation. Furthermore, Breuker et al. (2016) provide a technique with a global, post-hoc explanation by converting the learned model into a Petri net. By using a familiar representation for business users, they increase the comprehensibility of the process and the predictions made. Harl et al.

(2020) present a PPM technique based on graph neural networks that provides local, post-hoc explanations for single predictions and, to some extent, ante-hoc explanations by mapping the neural network structure to process graphs. Various other PPM techniques provide ante-hoc comprehensibility, such as Conforti et al. (2013), by using decision trees or Marquez-Chamorro et al. (2017), which present a technique based on decision rules.

*2.4. Evidence Sensitivity Analysis*

An approach for creating post-hoc explanations in BNs is evidence sensitivity analysis (ESA). ESA measures the sensitivity of posterior probabilities of a belief update to changes in the evidence (Kjaerulff and Madsen, 2007, p. 274). It provides answers to various questions that arise around reasoning under uncertainty with graphical models (e.g. BNs). For example, *which context attribute acts in favor of or against a hypothesis?*. This aids in the understanding of probabilistic inference in a BN and, therefore, to explain conclusions as well as decisions made under uncertainty by the network.

One possible application of ESA is to investigate the impact of different subsets of an evidence $E$ over a hypothesis $X = x$ (Kjaerulff and Madsen, 2007, p. 277 f.). In our case, the hypothesis is the event in a particular state $A = a$. It helps to identify context attributes, which act in favor of an event state, and variables which act against it. Kjaerulff and Madsen (2007) define the impact of a particular subset $E' \subseteq E$ on a certain state $x$ of a hypothesis variable $X$ by calculating the normalized likelihood (NL) of the hypothesis $x$ given the different subsets of the evidence $E'$. The NL for each subset acts as a measure for the impact of the particular subset on the hypothesis. If the NL of a subset $E'$ is above 1, the subset $E'$ acts in favor of the hypothesis. If the NL of a subset $E'$ is below 1, the subset $E'$ acts against the hypothesis.

## 3. Model Construction

This section focuses on the design of a context-sensitive technique for business process prediction based on a DBN. The technique consists of an offline phase for creating the model and an online phase, where the model is used to predict unfinished process instances, as shown in Figure 2.

*3.1. The Offline Phase*

Kjaerulff and Madsen (2007) described the model elicitation process for a DBN as a two-step process. In the first process step, the network structure is defined, and variables and their causal, functional, or informational relationships are identified. In the second step, initial probabilities are defined for the previously constructed structure (e.g., by randomly initializing them or by using known probabilities), followed by the learning of those probabilities from the data.

Defining the network structure can be done via a data-driven (automatic) process (e.g. Liang et al., 2020), manually (e.g. Uusitalo et al., 2018), or by a combination of both. In this work, the structure is manually specified to reflect our goal to account for cause and effect attributes. Therefore, we define the relation among cause and effect context and the process event variable. The parameters are then learned from data.

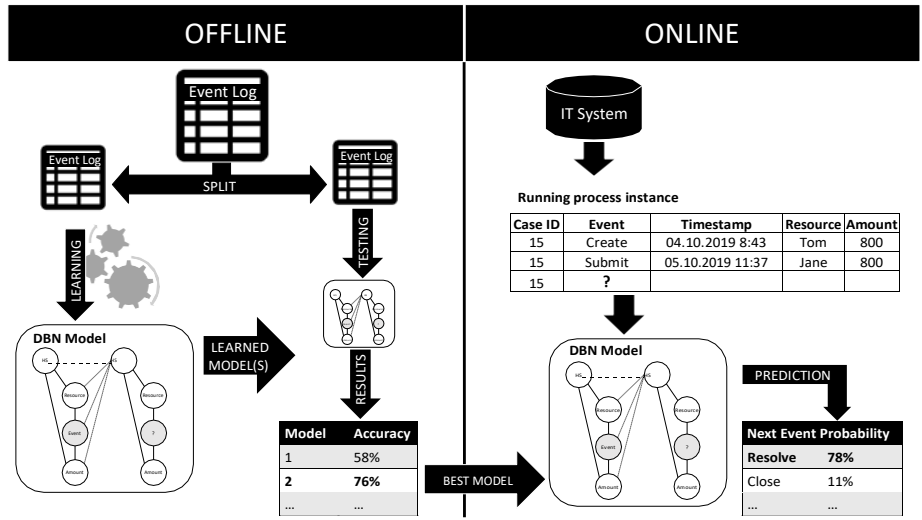

Figure 2: The CECA-DBN Architecture

According to Kjaerulff and Madsen (2007), graphical causal modeling requires three types of variables – problem variables, information variables, and mediating (or hidden state) variables – to be differentiated. Information variables can be further differentiated as background variables and symptom variables. Mediating variables, which are not observable, are used to increase the computing efficiency of the inference step by combining different information variables into one variable.

Problem variables represent the variables of interest. They are also referred to as hypothesis variables and are related to the predictions and decisions that are to be made. The posterior probability of such variables is usually computed based on observations on information variables.

Table 1 shows an example of an event log used in process mining. It reflects a sales process and consists of the mandatory unique instance identifier, an event label, a timestamp, and other attributes related to the process instance. In the prediction of process sequences, the problem variable is represented by the next unknown event, which is to be predicted.

Information variables provide information for solving the inference query on the problem variable. Information variables can be distinguished into two sub-categories: background variables and symptom variables. Symptom variables arise only after the observation of the problem variable, as they are a result of the observation. In contrast, background variables are information variables that are available before the occurrence of the problem variable (Kjaerulff and Madsen, 2007, p. 150). Therefore, they have a causal influence on problem variables, as well as the symptom variables, and can be the root of a probabilistic network. Related to context-sensitive PPM, background variables are consequently known in the time slice in which the prediction is made, but symptom variables are not. They are only available within the next time slice.

| Order | Event | Timestamp | Distance to Customer | Shipping Cost |
|---|---|---|---|---|
| 12345 | Receive Customer Order | 29.06.2020 19:39 | 500 km | - |
| 12345 | Calculate Shipping Route | 30.06.2020 11:24 | 500 km | - |
| 12345 | Contract Delivery Service | 30.06.2020 11:31 | 500 km | 5$ |
| 12345 | Package Goods | 05.07.2020 15:33 | 500 km | 6$ |

Table 1: Exemplary event log

For the event log shown in Table 1, a suitable example of a symptom variable is the *Shipping Cost*. Dependent on the event that is executed, the price has to be adjusted (e.g. because of packaging materials). The distance to the customer can be considered as a background variable, which influences the process execution (e.g. close customers will be delivered directly while distant customers require an external delivery service) but will not change based on the process flow.

In summary, when speaking about cause-effect relationships, context variables are differentiated in terms of variables that cause the next event and variables that are the effect of an event.

Fig. 3 shows the structure of a DBN that models a context-sensitive approach that considers context information and that differentiates between types of information variables. $H$ represents the hidden state of a particular time slice, as in the non-context-sensitive technique. The event of each time slice is depicted as $E$, while $B$ represents context information as a background variable, and $S$ represents context as a symptom variable.

The network structure consists of four types of intra-time-slice edges (i.e., relationships between the attributes of an event) per time slice and four types of inter-time-slice edges (i.e., relationships between the attributes of consecutive events). The four types of intra-time-slice edges are:

1. $H^{(t)} \rightarrow B^{(t)}$     Hidden state to background variable
2. $H^{(t)} \rightarrow E^{(t)}$     Hidden state to the event variable
3. $B^{(t)} \rightarrow E^{(t)}$     Background variable to the event variable
4. $E^{(t)} \rightarrow S^{(t)}$     Event variable to the symptom variable

Apart from their direct influence, variables also transmit information between variables that are not linked directly using three types of connections

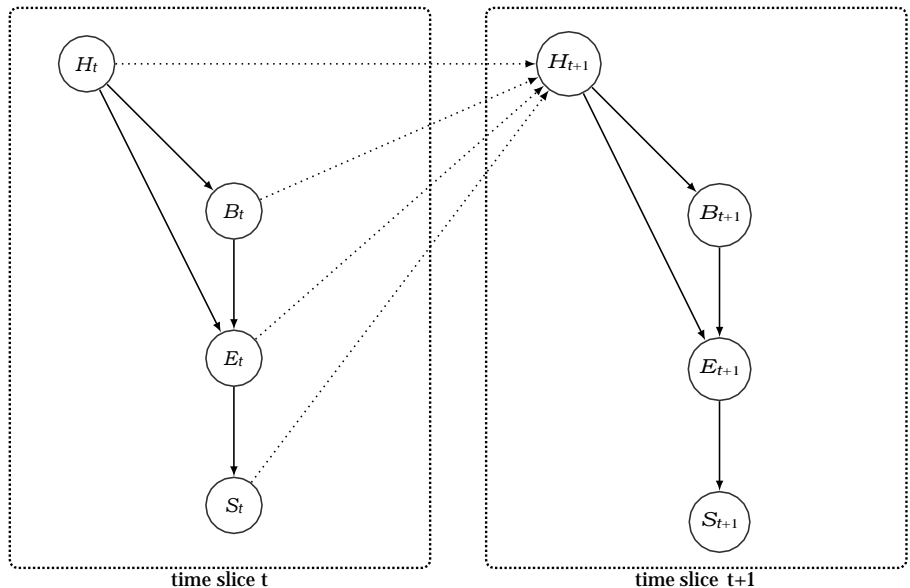

Figure 3: The Dynamic Bayesian Network Structure Representing the Generic Model

(Kjaerulff and Madsen, 2007): serial, converging, and diverging. Hence, the hidden state has an additional information flow to the event via the background variables as well as to the symptom variables (through the event) through a serial connection. The hidden state and the background variable have a converging connection through the event, hence competing to reason about the state of the event (inter-causal inference). The background information influences the symptom information via the event, a logical behavior in real-world settings. The four types of inter-time-slice edges are:

1. $H^{(t)} \rightarrow H^{(t+1)}$     Hidden state in $t$ to hidden state in $t+1$
2. $B^{(t)} \rightarrow H^{(t+1)}$     Background variable in $t$ to hidden state in $t+1$
3. $E^{(t)} \rightarrow H^{(t+1)}$     Event in $t$ to hidden state in $t+1$
4. $S^{(t)} \rightarrow H^{(t+1)}$     Symptom variable in $t$ to hidden state in $t+1$

The structure of the DBN is generic, i.e., independent of the domain. When it is applied to a specific domain, the domain specialist must define which variables will be considered as symptom and which will be considered as background information. Furthermore, in the current version of our model, we assume independence among the contextual information.

In the second process step, we follow Breuker et al. (2016) in initializing the parameters randomly, but we do not apply regularization. Again, we are following the design goal of creating a generic artifact, and we assume that no knowledge about the probability distributions of any of the attributes exists. We accept the risk of setting unfavorable starting parameters for the learning algorithm, which might lead to local optima or slow convergence but minimize the chance of finding local optima by rerunning the algorithm several times with different configurations (e.g., numbers of hidden states) (Breuker et al., 2016).

For the learning of the model parameters, we use a popular approximate learning algorithm for DBNs that Boyen and Koller (1999) developed, as exact inference is not feasible with a complex network structure and would prevent the applicability of our technique in practical settings. Boyen and Koller (1998, p.398) used an expectation maximization (EM) algorithm, "an iterative procedure that searches over the space of parameter vectors for one which is a local maximum of the likelihood function", and which is suitable for the problem at hand (Breuker et al., 2016).

### 3.2. The Online Phase

After learning the parameters for the network, predictions are made by means of an inference task on the event node $E^{(t+1)}$ in time slice $t + 1$ given evidence $E$. Table 2 depicts the two different ways context information is given as evidence $E$.

Table 2a presents the evidence when context information is modeled as a symptom variable. Here, the posterior probability of an event is computed as a result of the inference task: in this example, $P(E^{(4)} | E^{(3)} = D, S^{(3)} = z)$. In contrast to the symptom variables in Table 2a, Table 2b depicts evidence $E$ of context information modeled as a background variable. Since background information for time slice $t+1$ is known before the event $E^{(t+1)}$ is observed, the evidence $E$ contains observations for the time slice $t+1$, additional information that is also considered in the inference task. The inference query for the network structure with a background variable is $P(E^{(4)} | E^{(3)} = D, B^{(4)} = y, B^{(3)} = z)$.

Compared to the learning phase, the inference task is computationally inexpensive, so predictions can easily be made in real-time, an important feature for applying the technique in practice. Here, predictions should be available at the earliest possible point in time to give the best available support to decision-makers.

| (a) Context as Symptom Variable | | | | | (b) Context as Background Variable | | | | |
|---|---|---|---|---|---|---|---|---|---|
| Time slice $T$ | 1 | 2 | 3 | 4 | Time slice $T$ | 1 | 2 | 3 | 4 |
| Hidden State $H$ | | | | | Hidden State $H$ | | | | |
| Event $E$ | A | B | D | | Event $E$ | A | B | D | |
| Context $S$ | w | x | z | | Context $B$ | w | x | z | y |

Table 2: Exemplary Evidence *E* for the Introduced Structure

## 4. Implementation and Model Validation

We evaluated existing software packages for probabilistic models, such as GMTK [1], AmidstToolbox [2] and Bayes Net Toolbox [3] to implement the described model structure, called the Cause-Effect Context-Aware Dynamic Bayesian Network (CECA-DBN). The requirements to be met were to model custom DBNs network structures (i.e.,, the envisioned CECA-DBN structure) and to learn and make inferences on the model. We chose the Bayes Net Toolbox for MATLAB, as it is, to the best of our knowledge, the only package that fulfills all requirements. The source code of our implemented technique, the learned models as well as the data sets are available in a public Git repository [4].

To validate the model structure and the causal relations between the background variable and the activity as well as the activity and the symptom variable, we construct four synthetic data sets. The synthetic business process consists of four types of activities and two different process variants. Each data set includes a context variable with four possible values. Two of the data sets are constructed for a background variable structure, to validate the causal link between the background and activity variable, and two for a symptom variable based structure, to validate the link between the activity and symptom variable. For both variable structures, we construct one data set in which the context data describes a perfect causal relation with the activity (see Figure 3) and one data set with random context data that does not adhere to the causal relationships of background or symptom variables (i.e., the context attribute has no predictive value).

In all our experiments (on synthetic and real-world data sets), we split the data set into a 70-30 ratio of training and testing data to avoid overfitting. This approach simulates applying the technique to new, yet unseen data. We also apply 10-fold cross-validation to further improve the validity of our experiments and to ensure that we do not stumble on a model that performs exceptionally well by chance because the test data set was opportune (Kohavi, 1995). As quality metrics, we report accuracy and the F1-Score. We perform all calculations on a high-performance cluster, which supports fast and parallel execution of the various learning and inference tasks.

Table 3 shows the predictive performance of CECA-DBN given the four synthetic data sets. In the synthetic background structure, we predict the succeeding activity and in the symptom structure, we predict the succeeding symptom variable. CECA-DBN predicts the activity as well as the symptom variable perfectly in the two data sets where the causal relationship of the model is existent within the data. In the other two cases, the predictive quality for both quality metrics is much lower, even though the synthetic business process is not very complex. The perfect prediction of the data sets that include the causal relation between context attributes and the business process activity, which we model with CECA-DBN, validates the implementation and the underlying causal model.

---

[1] https://melodi.ee.washington.edu/gmtk/
[2] http://www.amidsttoolbox.com/
[3] https://github.com/bayesnet/bnt
[4] https://wiwi-gitlab.uni-muenster.de/j_brun17/ceca-dbn

| Structure        | Background |        | Symptom |        |
|------------------|------------|--------|---------|--------|
| Prediction Target | Activity  |        | Symptom |        |
| Causal Structure in Data | Yes | No | Yes | No |
| Accuracy         | 1          | 0,7374 | 1       | 0,2449 |
| F1-Score         | 1          | 0,6487 | 1       | 0,2935 |

Table 3: Quality Metrics for Model Validation on Synthetic Data

## 5. Experiment Design

To assess the technique, we benchmark it with various structures and input attributes against the PFA and an n-grams implementation. N-grams, which are popular in the domain of grammatical interference, provide frequency tables of word combinations or, in this instance, subsets of business processes. An n-gram counts the different variants of process executions and, given an instance's log, predicts the most probable next event based on the number of previous occurrences (Verwer et al., 2014). The $n$ in n-grams stands for the length of the fragments that the method works on. In our case, we performed an evaluation of 3- to 7-grams for each experiment and used the best performing one as a baseline for comparison (see Appendix Table 7 and 8).

The PFA serves as an additional baseline for our technique, as it represents the same structure in the absence of context. The n-grams is a useful baseline from which to evaluate whether a log has many similar process flows, so the next event might be straightforward to predict. CECA-DBN's performing well in a specific log while n-grams do not indicate an actual improvement. CECA-DBN also outperforming the PFA can show that it makes sense to include context and, more importantly, to model context in regards to cause and effect. Besides, we compare our measures with state-of-the-art publications in the domain of PPM.

We use five established event logs for our analysis. A de-facto standard repository for business process event logs in the domain stems from the Business Processing Intelligence (BPI) Challenges that have been held since 2011. For our purpose, we selected the event logs from 2012 and 2013, which fit the purpose of predicting business process sequences the best and have been used in previous works most often (e.g., Breuker et al. (2016); Tax et al. (2017); Tama and Comuzzi (2019); Camargo et al. (2019)), which serves the comparability of our work.

The *BPI 2012* event log is taken from the loan application process of a Dutch financial institute (Van Dongen, B.F. (Boudewijn), 2012). The overall process is subdivided into three sub-processes: offer, application, and work item sub-processes. The respective log files are denoted *BPI2012o, BPI2012a and*

*BPI2012w* accordingly. Apart from the process instance identifier, the event, and the time stamp, the logs include the contextual attributes of *Resource* (handling employee) and the *Requested Loan Amount*, which could include valuable information on possible next events. The work item process is controlled by a human that interacts with an IT system. The offer and application subsets are considerably smaller and not generated by human interaction; instead, they automatically log the application or order.

We also use the *BPI 2013* event log, which was provided by Volvo IT Belgium and includes the log of an incident and a problem management system called VINST (Steeman, 2013). The log is split into sub-logs: Problems (*BPI2013p*) and Incidents (*BPI2013i*). For our evaluation, we selected the attributes *Involved ST Function Div*, which represents the division of the incident handling support team, and the *Status*, which is the parent category of the process activities. The log contains various additional attributes, such as *Product* or *Owner Country*. However, most of them are related to the overall process instance, so they may not have much direct influence on the next event. Here, the *Problem* sub log is considerably smaller than the *Incident* sub log.

| Event log | #Instances | Lengths | Events | #Events | Context Attributes [type, values] |
| --- | --- | --- | --- | --- | --- |
| BPI2012a | 13.087 | 3-8 | 10 | 60.849 | Resource [integer, 112-11.339] Amount [integer, 0-99999] |
| BPI2012o | 5.015 | 3-30 | 7 | 31.244 | Resource [integer, 112-11.339] Amount [integer, 25-99000] |
| BPI2012w | 9.658 | 1-74 | 6 | 72.413 | Resource [integer, 112 - 11.339] Amount [integer, 0-99999] |
| BPI2013i | 7.554 | 1-123 | 13 | 65.533 | Status [String with 4 values] Function [String with 24 values] |
| BPI2013p | 1.487 | 1-35 | 7 | 6.660 | Status [String with 4 values] Function [String with 24 values] |

Table 4: Characteristics of the Used Data Sets

Table 4 gives an overview of the chosen logs, their attributes, and their characteristics. We did some minimal preprocessing to the data sets for our initial implementation. We excluded traces that had fewer than three events, as three is the minimum number of events that are necessary for the inference task (because of the two-time slices DBN and the third event to be predicted). We also discretized the numeric *Resource* and *Amount* attributes of the *BPI2012* data set into forty intervals, as otherwise, the model would become too complex.

Following the model construction process in section 3, we construct two example instantiations of CECA-DBN to illustrate the process and the subsequent evaluation. In the first step, a process analyst familiar with the business process and its possible cause and effect relations with the surrounding context variables needs to decide on the context attribute to include and its causal relationship (i.e., background or symptom) to the activity.

Confronted with the BPI2012 event log and its underlying process, the process analyst believes that the *resource* attribute is a suitable *symptom* context variable because depending on the event that was executed, a future handling *resource* could be assigned. He or she, therefore, believes that including it can improve the predictive quality of the model.

> **Structure Hypothesis 1 (SH1):** Including the resource attribute of the BPI2012 data set as a symptom context variable in CECA-DBN improves the predictive quality.

In the case of the BPI2013 data set, the process analyst believes that the *status* attribute stands in a cause-relation to the activity, since the activity is a sub-status of the overall *status*, and should, therefore, be included as a *background* context variable.

> **Structure Hypothesis 2 (SH2):** Including the status attribute of the BPI2013 data set as a background context variable in CECA-DBN improves the predictive quality.

## 6. Results and Evaluation

As mentioned in section 3, a criticial step of the offline phase is the specification of the background and symptom variables by the process analyst. Along with the instantiations of CECA-DBN to address **SH1** and **SH2**, we also ran experiments with further configurations of the available context attributes of the five data sets. We used each context attribute, as background and symptom variables, to compare results and investigate the appropriateness of the design choice. We also learned a PFA model without context attributes and a number of n-gram models on the same data sets. Both serve as baselines in our evaluation. The quality metrics of the 10-fold cross validation for all techniques, structures, and logs are listed in Table 5.

A look at the baseline results shows that our implementation of the PFA outperforms the n-grams approach in every metric except the *F1-Score* for *BPI2012a* – a result that is in line with previous research (Breuker et al., 2016). For each log, except *BPI2012o* and the *F1-Score* of *BPI2012W*, at least one configuration of CECA-DBN outperforms the PFA in *Accuracy* and *F1-Score*. That one context-sensitive structure nearly always outperforms the PFA supports the results of previous research (Schönig et al., 2018; Evermann et al., 2017), and our understanding that the inclusion of context is a feasible way to improve business process predictions. Most important, it shows that we can configure CECA-DBN in nearly all test cases to outperform the established PFA structure.

However, in most cases, including the context attribute does not improve prediction *Accuracy* but results in a small decline. This trend makes sense, as not all of an event log's attributes – and not all of those used for the benchmark – contain valuable information regarding the next process event. However, their inclusion in the model makes the DBN structure more complex, so learning and performing accurate inference on these models are also more complex. Hence, it is important to evaluate which attribute to include in the model but also to think carefully about their causal relationship to the event (i.e.,, cause or effect).

| BPI2012a | Context | Amount | | Resource | | - | - |
|---|---|---|---|---|---|---|---|
| | **Structure** | Sympt. | Backg. | Sympt. | Backg. | PFA | n-gram[1] |
| | **Accuracy** | **0,6769**[2] | 0,5319 | 0,6532 | 0,6541 | 0,6580 | 0,5764 |
| | **F1-Score** | 0,6507 | 0,4541 | 0,6818 | 0,6856 | 0,7059 | **0,7281** |
| BPI2012o | **Context** | Amount | | Resource | | - | - |
| | **Structure** | Sympt. | Backg. | Sympt. | Backg. | PFA | n-gram[1] |
| | **Accuracy** | 0,78 | 0,685 | 0,7387 | 0,7749 | **0,7845** | 0,6997 |
| | **F1-Score** | 0,7101 | 0,5518 | 0,6826 | 0,6517 | **0,7552** | 0,7179 |
| BPI2012w | **Context** | Amount | | Resource | | - | - |
| | **Structure** | Sympt. | Backg. | Sympt. | Backg. | PFA | n-gram[1] |
| | **Accuracy** | 0,7402 | 0,6487 | **0,8592** | 0,8147 | 0,7440 | 0,7331 |
| | **F1-Score** | 0,5493 | 0,4086 | 0,6695 | 0,6451 | **0,6806** | 0,5782 |
| BPI2013i | **Context** | Function | | Status | | - | - |
| | **Structure** | Sympt. | Backg. | Sympt. | Backg. | PFA | n-gram[1] |
| | **Accuracy** | 0,5966 | 0,5540 | 0,5840 | **0,8588** | 0,5875 | 0,5140 |
| | **F1-Score** | 0,4925 | 0,4182 | 0,5306 | **0,8084** | 0,5643 | 0,3116 |
| BPI2013p | **Context** | Function | | Status | | - | - |
| | **Structure** | Sympt. | Backg. | Sympt. | Backg. | PFA | n-gram[1] |
| | **Accuracy** | 0,5607 | 0,3997 | 0,5817 | **0,8397** | 0,5744 | 0,4877 |
| | **F1-Score** | 0,3993 | 0,305 | 0,5684 | **0,7351** | 0,5452 | 0,3749 |

[1] Best value of 3- to 7-gram.
[2] The best (highest) value of each metric (row) is marked in bold.

Table 5: Classification Metrics of all Logs and Structures

In several configurations, the context-sensitive technique improves the predictive model considerably in all metrics. One of these cases is the *BPI2012w* log with the *Resource* attribute in a symptom structure as well as background structure. Here, the symptom structure performs best in *Accuracy*, while the PFA is superior in regards to the *F1-Score*. In *BPI2012a* and *BPI2012o*, we do not observe an improvement in predictive quality by including the *resource* attribute as a background variable. Therefore, we need to reject **SH1**, and alternative structures and inputs might need to be considered by the process analyst in further iterations.

Out of the *BPI2012* data, only the *BPI2012w* log is created through direct human interaction. The *Resource* attribute represents a processing unit, such as an employee or functional division, so the relationship between the log and the resource possibly explains why the context-sensitive technique performs best only in the case of *BPI2012w*. The *Resource* usually only changes once or twice in a process instance. The metrics indicate that, in the case of human interaction (*BPI2013w*), the context attribute, modeled as a symptom variable, includes valuable information for predicting the next event, possibly because of this dynamic change in the attribute. However, further investigation of this phenomenon is required to validate this hypothesis.

Two more cases in which CECA-DBN significantly outperforms the baseline approaches occur when the *Status* attribute is included as a background variable in the *BPI2013i* and *BPI2013p* logs. This example shows how a context attribute that is known before the event (i.e., background) can have a substantial impact on the event and improve the predictive model. Here, the *Status* attribute represents a grouping of the sub-status values, which are the process instance events. Therefore, it is reasonable that this information is valuable in a background structure but not in a symptom structure, which the quality metrics verify. In contrast, the symptom structure performs similarly to the PFA in both cases. Therefore, we can accept **SH2**.

To better comprehend the results, we additionally performed an ESA for the configurations of **SH1** and **SH2** to investigate the impact of the context attributes on the prediction probabilities of the correct next event.

Fig. 4 visualizes the results of the ESA in the form of two box plots. The analysis was conducted on each trace of the event logs, and the NL value is shown on the y-axis. A NL value lower than one means that the evidence (context information) has a negative impact and a value higher than one a positive impact on the prediction probability of the correct next event in the trace. For **SH1**, the quartile values range from 1,01 to 1,32, with a median of 1,04, which is in line with the results of our quality metrics. The inclusion of the context information does not have a significant effect on the prediction probabilities since the vast majority of values is situated close to 1 (having no effect). The values of **SH2** show a different picture. Here, the quartile values range from 1,82 to 4712,88, with a median of 3,83, which shows that the inclusion of the status context attribute mostly has a strong positive influence on the prediction probability of the correct next event. The ESA thereby confirms that the high predictive quality of **SH2** is caused by the inclusion of the context

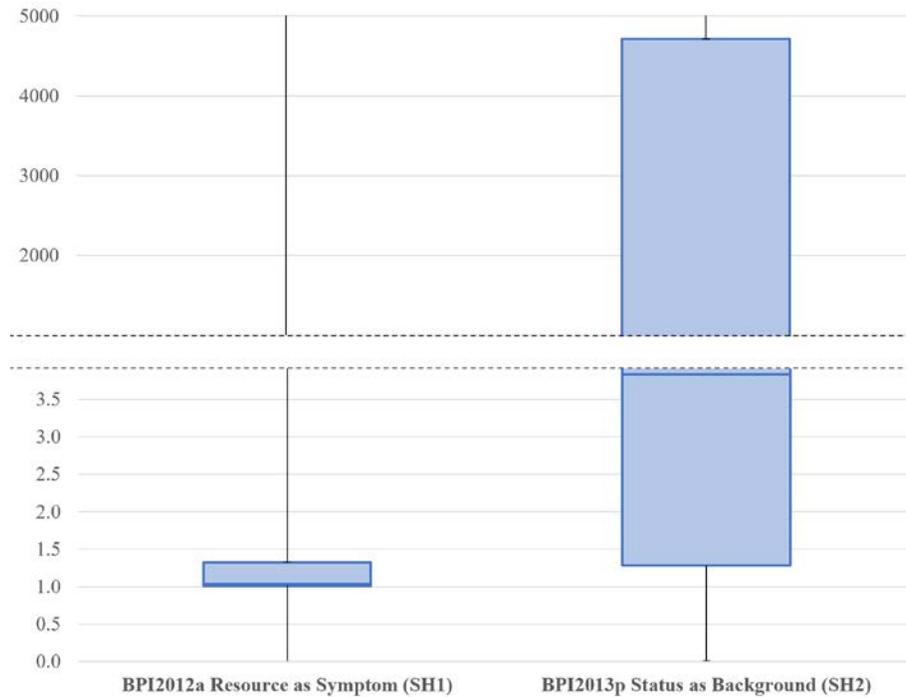

Figure 4: Evidence Sensitivity Analysis for Structure Hypothesis 1 and 2

information.

CECA-DBN performs similarly to or sometimes better than other prediction techniques that were computed on the selected data sets. Compared to Breuker et al. (2016), we report higher *accuracy* values in four out of five cases, and CECA-DBN outperforms all of Tama and Comuzzi's (2019) classification techniques for the *BPI2012* and the *BPI2013* logs. For the *BPI2012w* data set, Tax et al. (2017) achieved an accuracy of 76 percent with their technique based on LSTM neural networks without considering context and Camargo et al. (2019) report 77,8 % extending the LSTM architecture by context. CECA-DBN clearly outperforms both with an accuracy of 85.9 % when the *resource* attribute is modeled as a symptom variable. Table 6 summarizes these correlation-based comparisons, which, however, need to be considered with caution since the pre-processing, sampling of data, or the implementation of quality metrics can differ.

In summary, the results CECA-DBN achieved show that it can compete well with state-of-the-art techniques as well as simpler techniques like a PFA and n-grams. More importantly, the results show that including context is not guaranteed to improve the prediction quality, but the influence of each variable on the activity must be modeled adequately.

| Log | Technique | Author | Accuracy | F1-Score |
|---|---|---|---|---|
| BPI2012a | CECA-DBN | *This technique* | 0,6769 | 0,6856 |
| | RegPFA | Breuker et al. | 0,801 | - |
| | Multi-stage deep learning | Mehdiyev et al. | 0,824 | - |
| BPI2012o | CECA-DBN | *This technique* | 0,78 | 0,7101 |
| | RegPFA | Breuker et al. | 0,811 | - |
| | Multi-stage deep learning | Mehdiyev et al. | 0,821 | - |
| BPI2012w | CECA-DBN | *This technique* | 0,8592 | 0,6695 |
| | RegPFA | Breuker et al. | 0,719 | - |
| | LSTM | Camargo et al. | 0,778 | - |
| | LSTM | Tax et al. | 0,76 | - |
| | Multi-stage deep learning | Mehdiyev et al. | 0,831 | - |
| BPI2012* | LSTM | Camargo et al. | 0,786 | - |
| | Credal Decision Tree | Tama and Commuzzi | 0,56 | - |
| BPI2013i | CECA-DBN | *This technique* | 0,8588 | 0,8084 |
| | RegPFA | Breuker et al. | 0,714 | - |
| | Random Forest | Tama and Commuzzi | 0,69 | - |
| | Multi-stage deep learning | Mehdiyev et al. | 0,663 | - |
| BPI2013p | CECA-DBN | *This technique* | 0,8397 | 0,7351 |
| | RegPFA | Breuker et al. | 0,69 | - |
| | Multi-stage deep learning | Mehdiyev et al. | 0,782 | - |

[*]These results were computed on the merge of the above listed BPI2012 data sets.

Table 6: Comparison of Prediction Technique Performances

## 7. Discussion

Our work contributes to the PPM domain by introducing a new technique for next event prediction based on DBNs that includes context attributes and the process flow into its models.

We model context attributes as either cause variables or effect variables, an approach that was shown to be reasonable by means of an evaluation with several real-life data sets. The results demonstrate that including context does not always improve prediction accuracy, but the causal relationship between an attribute and the event must be correctly indicated. We hope this work stimulates a discussion about different properties, structures, and kinds of event log attributes and how they can contribute to PPM.

The comprehensibility of ML techniques has been found important for user adoption. We present a technique based on probabilistic models that provides more comprehensibility by enabling post-hoc explanations through access to the CPDs. We showed that the technique can compete with state-of-the-art techniques like recurrent neural networks. The structure behind CECA-DBN is easily comprehensible, and it could be adapted by, for example, domain experts to introduce their knowledge about the business process (Swartout et al., 1991). End users are valuable providers of data, and their understanding of the data and the process should be exploited for more accurate predictions. We instantiated the CECA-DBN artifact so that it can be applied by practitioners and researchers.

A limitation of our work is indicated by the benchmark results, which show that the data sets might not have been ideal for our use case. However, they are the de facto standard in the PPM community, and their use enables us to compare our results to other works in the field. We reported two quality metrics for the multi-class classification problem but did not perform a detailed per-case analysis of each log, structure, and metric. In future work, additional analyses, such as considering the per-class non-macro averaged metrics, of the various experiments could potentially provide further insights.

Most importantly, we did not evaluate our technique including both a symptom variable and a background variable at the same time in favor of clearly differentiating between the impact of each configuration on the prediction's quality. We acknowledge that including both kinds of variables might have unforeseen effects that should be subject to further investigation.

We plan to address these limitations by applying CECA-DBN to other data sets. Context information stems not only from the systems that generate the event logs but can also come from attached systems or even unrelated systems like weather stations. Incorporating such information into the model is a reasonable next step and should provide ample opportunities to improve the predictive power of CECA-DBN models.

A discussion of use cases and the objectives of predictive techniques in the field of PPM is required. Research from other fields that are developing ML techniques has shown that developing a one-size-fits-all solution is rarely the right choice (Erickson et al., 2018). Techniques should be tailored to specific tasks, such as detecting non-compliant, expensive, risky, or costly events. For such techniques, reporting (per-class) quality metrics like precision and recall would deliver valuable insights that we do not see currently.

In this work, we assumed independence among the contextual information. One relevant future investigation is if explicitly modeling the dependency among contextual information brings benefits for the prediction.

Furthermore, a promising path for future research could be to use our technique for process analysis. Marquez-Chamorro et al. (2017) state that the decision rules used by their technique for the prediction could deliver insights about the process to users. Similarly, our technique could be used to understand better both the root causes for process performance and the impact of actions taken during process execution.

Finally, we have demonstrated the comprehensibility property of CECA-DBN by creating post-hoc explanations, but we have not included the explanation as part of our technique. Visualizing how evidence impacts the possible execution flow of a process instance could be an exciting path to extend our technique in the future (Champion and Elkan, 2017).

## 8. Conclusion

We designed a PPM technique (CECA-DBN) that is both process-aware and context-sensitive based on established research from the area of DBNs. Our differentiation of types of context based on them having a cause (background) or effect (symptom) relationship to the process flow, is novel in the PPM field. Through our benchmark on established data sets, we showed that CECA-DBN can improve the predictive quality of probabilistic models by including additional context information.

We encourage future research to follow our proposition and to investigate further the effects of context attributes on and their relationship to the process flow of business processes.

## Acknowledgments

The research leading to these results received funding from the European Union's Horizon 2020 research and innovation program under the Marie Skłodowska-Curie grant agreement No. 645751 (RISEBPM). The fourth author received a grant from Österreichische Akademie der Wissenschaften.

**Appendix**

| n-gram | n=3 | n=4 | n=5 | n=6 | n=7 |
|---|---|---|---|---|---|
| **BPI2012a** | 0,3599 | 0,3599 | 0,5044 | 0,5348 | 0,5764 |
| **BPI2012o** | 0,4411 | 0,5469 | 0,6154 | 0,6604 | 0,6997 |
| **BPI2012w** | 0,7331 | 0,7302 | 0,7306 | 0,7241 | 0,7231 |
| **BPI2013i** | 0,2484 | 0,4110 | 0,4804 | 0,5050 | 0,5140 |
| **BPI2013p** | 0,4735 | 0,4758 | 0,4853 | 0,4877 | 0,4860 |
| **Average** | 0,4512 | 0,5048 | 0,5632 | 0,5824 | 0,5998 |

Table 7: N-gram Accuracy Benchmark

| n-gram | n=3 | n=4 | n=5 | n=6 | n=7 |
|---|---|---|---|---|---|
| **BPI2012a** | 0,5515 | 0,5515 | 0,7281 | 0,6550 | 0,6472 |
| **BPI2012o** | 0,5780 | 0,6003 | 0,6112 | 0,6884 | 0,7179 |
| **BPI2012w** | 0,5782 | 0,5386 | 0,56 | 0,558 | 0,5519 |
| **BPI2013i** | 0,1371 | 0,2612 | 0,2873 | 0,3063 | 0,3116 |
| **BPI2013p** | 0,3627 | 0,3749 | 0,3621 | 0,3677 | 0,3660 |
| **Average** | 0,4415 | 0,4653 | 0,5097 | 0,5151 | 0,5189 |

Table 8: N-gram F1-Score Benchmark